\documentclass[11pt]{article}
\usepackage{acl2016}
\usepackage{times}
\usepackage{url}
\usepackage{latexsym}
\usepackage{color}

\usepackage{amsmath}
\usepackage{amssymb}
\usepackage{amsfonts}
\usepackage{tabu}
\usepackage{xcolor}
\usepackage{graphicx}
\usepackage{bm}
\usepackage{multirow}

\aclfinalcopy % Uncomment this line for the final submission
\title{Sense Perception Common Sense Relationships}

\author{Ndapa Nakashole \\
	Computer Science and Engineering  \\
	University of California, San Diego\\
	La Jolla, CA 92093 \\
	{\tt nnakashole@eng.ucsd.edu} \\}

\date{}

\begin{document}
	\maketitle
	\begin{abstract}
%		Knowledge bases  contain  relationships
%		about named entities such as people, locations, and organizations.
		 Often missing in existing knowledge bases of facts, are relationships
		that encode common sense knowledge about  unnamed entities. In this paper, we propose to extract novel, common sense relationships pertaining to sense perception concepts such as sound and smell. 
%		A  limiting factor to learning these kinds of relationships is the  difficulty of obtaining labeled data. In this paper,  we show that by exploiting large corpora we can generate  easy yes/no crowd-sourcing  questionnaires to obtain labeled data . We show that the resulting data can be used to train highly accurate models as shown by our experiments with both linear models, and non-linear neural network models. 
	\end{abstract}

	\section{Introduction}
	%{\color{red} Yes or no answers. \textit{``yes/no/notsure"} }
%	Knowledge acquisition efforts seek to produce assertions
%	and facts that support inference, question answering, and reasoning \cite{havasi2007conceptnet,DBLP:conf/aaai/TandonMW11,Bollacker2008,hoffartetal2012a,MitchellCHTBCMG15}. 
	%Existing knowledge base construction methods have successfully produced factual knowledge about named entities  such as people, and locations \cite{Bollacker2008,hoffartetal2012a,MitchellCHTBCMG15}. Common sense, on the other hand, has been much more difficult to produce and success has mostly been limited to part-whole relations (i.e. a car has wheels), and taxonomic  relations (i.e., a dog is an animal) \cite{havasi2007conceptnet,DBLP:conf/aaai/TandonMW11}. 
	%Even  crowd-sourced common sense knowledge bases, spanning decades of manual effort,  for example ConceptNet~\cite{havasi2007conceptnet}, are still sparsely  populated.
	% %as we show in our experiments. 
	%\paragraph{Goal. }
We seek to extract novel common sense relationships, with a focus on   concepts that are discernible by sense, for example, sound and smell.  
	% For example,  extract sound sources (i.e. dogs bark, birds chirp) or extract the sentiment of smells ( i.e., smell of lavender  is  pleasant, whereas the smell of  rotten eggs  is unpleasant).
	There are various natural language understanding tasks where this type of knowledge is useful: consider the  problem of co-reference resolution as it occurs in the following sentences:  (s1.) As the cat approached the dog, it started \textit{barking} furiously; (s2.) As the cat approached the dog, it started \textit{meowing} furiously.   We can easily determine  that in s1, the pronoun ``it'' refers to the dog, whereas in s2, ``it'' refers to the cat. However, for a machine reading method to  correctly resolve co-reference in s1) and s2), it  requires access to background knowledge that asserts  that barking and meowing are sounds produced by dogs and cats, respectively. This type of knowledge is what we aim to extract in this paper.
	One of the factors impeding progress in common sense knowledge acquisition is  the lack of labeled data. Prior work has shown that it  can be straightforward  to obtain training data for identifying  relationships between named  entities such as companies and their headquarters, or  people and their birth places \cite{havasi2007conceptnet,DBLP:conf/aaai/TandonMW11,Bollacker2008,hoffartetal2012a,MitchellCHTBCMG15}.  Examples of  such relationships can be found in semi-structured  formats on the Web\cite{DBLP:conf/www/WuW08,DBLP:conf/icdm/WangC08}.  This is not the case for common sense relationships. 
	%Fist, we would like to recognize  mentions of sensory perception concepts. For example, classify  ``screeching tires" and  ``birds chirping" as mentions of audible concepts, and   ``perfume" and `` burning rubber"  as  mentions of olfactible concepts.   It is worth noting that we are recognizing \textit{mentions} of  sensory perception concepts. We do not perform
	%co-reference resolution. Therefore,
	%our setting resembles the established task of entity recognition\cite{DBLP:conf/acl/FinkelGM05,DBLP:conf/conll/RatinovR09}, with the difference being that we do not work on named entities, but focus on the case of  un-named entities where some of the features useful for named entity recognition do not apply.
	% Second, we would like to  extract relations pertaining to senses. 	
	%\end{enumerate}
	%\paragraph{Contribution.}
	Our contributions in this work are three-fold. First, we propose to extract novel relationships commonly absent in existing knowledge bases. Second, we propose a method for generating labeled data  by leveraging large corpora and   yes/no  crowd-sourcing questionnaires. Third, using the resulting labeled data, we train both a  linear model and memory neural network models, obtaining high accuracy on the task of extracting these  previously under-explored relationships.
	To focus our task, we consider three relations pertaining to sense perception of  sound and smell.
	Namely: 1) \textit{soundSourceRelation}, 2) \textit{soundSceneRelation}, and 3) \textit{smellSentimentRelation}.
	
	\section{Sound-Source Relationship}
	The sound-source relationship represents information about which objects produce which sounds. For example that planes and birds are capable of \textit{flying},  the wind  \textit{blows}, and   geckos \textit{bark}. Obtaining sufficient labeled data to learn an extractor for this relationship is non-trivial, we propose one approach in the next section. 
	%One of the factors impeding
	%progress in common sense knowledge acquisition
	%is the lack of labeled data.
	%In this section,  we present an approach for obtaining labeled data for the sound-source relationship.
	
	%{\color{red}(Describe one of the factors impeding progress in acquiring common sense knowledge.  Show our approach for obtaining training data, show the figure about labeling sounds in text. We trained a sequence classifier, which tags names, entities, etc.) One of the factors impeding
	%progress in common sense knowledge acquisition
	%is the lack of training data.}
	\subsection{Labeled Data Generation} \label{labsoundsources}
	One option for obtaining labeled data is to do a cold call on a crowd-sourcing platform by asking crowd workers to list examples of sounds and their sources. However,  such an approach  requires crowd workers to  think of examples without clues or memory triggers.  This   is time consuming and  error prone.  Additionally, this means that the  monetary cost  could be substantial. We propose to  exploit a large corpus to obtain preliminary labeled data. This enables us to only need    crowd workers to filter the data through a series of  \textit{``yes/no/notsure"}    questions. These type of questions require little effort  from crowd workers while mitigating the amount of noisy input that one could get from open-ended,   cold call,  type of questions.

	To pose filters  to crowd workers in the form of  \textit{``yes/no/notsure"} questions, we need a list of plausible sound-source pairs.  To this end, we propose a lightly supervised corpus-based technique. First,  we  identify which  phrases refer to sounds using a high yield, but potentially noisy pattern. In particular, we  apply the following pattern  to a large corpus \footnote{In our experiments, we used the English part of  ClueWeb09; http://lemurproject.org/clueweb09/}:
	``\text{ sound} \text{ of } \text{ \textless y\textgreater}". The result  is a large  collection of  occurrences such as:
	%\begin{eqnarray*}
	``\text{ sound} \text{ of \textit { singing children}}". This step produced a list of 134,471 unique  phrases that potentially refer to sounds.   To evaluate accuracy, we randomly selected a sample of  500  phrases   and asked 3 crowd workers  per  phrase, on Mechanical Turk,  to say \textit{``yes/no/notsure"} if they agree the  phrase refers to a sound concept.  By  majority vote measure, $73.4\%$ of the 500  phrases where considered true mentions of sounds, with a moderate agreement rate of 0.51 Fleiss $\kappa$. 
	
%	\begin {table}[t]
%	\centering
%	\begin{tabular}{ll}
%		\hline
%		%plane flying & flying bird\\
%		rattling chains & shattering pottery \\
%		howling wolves & 	skidding tires \\
%		%	bouncing balls   & wind whistling \\
%		%	squealing brakes & blowing wind \\
%		%	cork popping  & cat meowing \\
%		%	water splashing & 	jingling bells \\
%		%	people yelling & revving engines \\
%		%	people laughing & yapping dog  \\
%		bickering children& panting dogs \\
%		singing children & 	honking cars\\
%		squawking bird & barking geckos\\
%		%twittering birds & shrieking children \\
%		croaking frogs & cat purring\\
%		%tearing cloth & door sliding \\
%		
%		\hline
%	\end{tabular}
%	\caption{Example phrases which mention sounds}
%	% These phrases can be used to generate a list of plausible sound-source to be filtered by crowd-sourcing annotation.
%	\label{tab:verbnounexamples}
%	\end {table}
	
	This annotation result indicates that a substantial number  of the phrases generated by the pattern indeed refer to sound concepts. We  therefore use these phrases to generate a list of plausible sound-source pairs.
	%Given that the accuracy of pattern produces a large number of  phrases that refer to sound concepts, we next used these  phrases to generate a list of plausible sound-source pairs.
	% Notice that this list does not need to be perfectly accurate as it will be annotated through crowd-sourcing.
	One important observation we made was that  about  20,000~(15\%) of the  134,471  phrases  are bi-grams of the form:  ``verb noun" or ``noun verb" where  in both cases, the verb is in the  gerund or present participle V-\textit{ing} form.  For example, \textit{birds chirping, cars honking,squealing brakes}, etc.  From  phrases of this kind, we  create verb-noun pairs, that we treat as plausible sound-source pairs where the verb is the \textit{sound} and the noun is the \textit{source}.  
%	We  thus present these pairs to Mechanical Turk for annotation. In the  Mechanical Turk task, we present the verb-noun pair within a context of a sentence which mentions pair. 
	We then asked crowd-workers to decide if the source (noun) produces the sound (verb). Thus from ``birds chirping"   we generate the question, ``Is  \textit{chirping} a sound produced by  \textit{birds}?";  Negative examples  include: ``surrounding nature", and ``Standing ovation", i.e., standing is not a sound made by ovation.
	  We generated 634 such questions on which we obtained a moderate  inter-annotator agreement rate of Fleiss $\kappa =0.57$, see Table \ref{tbl:soundsoucesMturk}. We use the resulting  labeled data to train two types of learning methods.
	
	\begin{table}[t]
		\centering
		\begin{tabular}{|l|r|}
			\hline
			& $Fleiss$ $\kappa$\\
			\hline
			%
			%soundSource & 45.27\% & 54.42\% &  0.57\\ \hline
			%smellSentiment &  38.83\% & 39.33\%   & 0.43\\ \hline
			soundSource &  0.57\\ \hline
			soundEnvironment  & 0.35\\  \hline
			smellSentiment &  0.43\\ \hline
		\end{tabular}
		\caption{Fleiss  $\kappa$. inter-annotator agreement rates for the three relations on yes/no type crowd-sourcing tasks.
		}
		\label{tbl:soundsoucesMturk}
	\end{table}

	%\subsection{Learning Methods}  
	
	\subsection{Linear Learning Model}\label{linearmodel} 
	The learning problem for the sound-source relationship is as follows: given a  bi-gram phrase $n$ of the form ``verb noun" or ``noun verb", we wish to classify yes or no if a given noun, denoted by $w_{src}$, produces the verb, denoted by word  $w_{snd}$, as a sound.  As a linear solution to this problem, we train a logistic regression classifier. The features we use are the vectors representing the word embeddings of  $w_{src}$ and $w_{snd}$, denoted by $\bm{v}_{src}$,   and    $\bm{v}_{snd}$. In our experiments, we use  the 300-dimensional Google News pre-trained embeddings \footnote{https://code.google.com/archive/p/word2vec/}. There are several ways in which we combine $\bm{v}_{src}$,   and    $\bm{v}_{snd}$ into a single feature vector:\\
%	\begin{description}
		%\vspace{-0.4cm}
	\textbf{Vector Concatenation:}
		$v=  concat(\bm{v}_{src}, \bm{v}_{snd})$\\  Size of $v$, $|v| = |\bm{v}_{src}|$ + $|\bm{v}_{snd}|$\\
	\textbf{	LSTM encoder} : $v= lstm( \bm{v}_{src},\bm{v}_{snd})$ \\
		An LSTM \cite{DBLP:journals/neco/HochreiterS97a} recurrent neural network is used to encode the  phrase containing $\bm{v}_{src}$  and    $\bm{v}_{snd}$.
		$|v|= h$, where $h$ is the hidden layer size of the neural network.  \\
	\textbf{	Source minus sound}: $v= \bm{v}_{src} - \bm{v}_{snd}$\\
		$|v| = |\bm{v}_{src}| = |\bm{v}_{snd}|$\\
		\textbf{Sound minus source}: $v= \bm{v}_{snd} -  \bm{v}_{src}$\\
		$|v| = |\bm{v}_{src}| = |\bm{v}_{snd}|$

	\subsection{Memory Networks Learning  Model}\label{memorynetworks}
	In addition to the variations of the linear model, we also trained a non-linear model in the form of  memory networks  \cite{DBLP:conf/nips/SukhbaatarSWF15}.
	Memory networks have been recently introduced, they  combine their inference component  with a memory component.  The memory component serves as a knowledge base or history  vault  to recall words or facts from the past. For the task of
	relation extraction, the memory network model  learns a scoring function to rank relevant memories (words) with respect to how much they  express a  given relationship. This is done   for  a given argument pair  as a query, i.e.,  a sound-source pair. At prediction
	time, the model finds \textit{k} relevant memories (words) according to the scoring function and conditions its output on these memories.  In our experiments, we explore different values of  $k$, effectively changing how many memories (words), the model conditions on. We  report results for up to $k=3$ as we did not see improvements for larger values of $k$. 
	
	%{\color{red} {What is the input, what is the output.}  }

	%With memory networks, we wanted to explore a different non-linear model. A crucial component with memory networks is that they combine an inference components combined with a long-term memory component; they learn how to use these jointly.  We treat the noun phrase $n$ as the query, and the memories are the words in the sentence, this way we can see, by emplying attention, if any part of the sentence tells us or gives us a clue about the sound source relationship. Memory networks hop over the memories, a variable number of times, we specified different hops values.  
	%Next, we present the results of training the memory networks model, and the linear model using the labeled data generated by our approach as described in section~\ref{labsoundsources}.
	
	\subsection{Sound-Source Evaluation } 
	Both the linear model and the memory networks models were implemented using Tensorflow.
	For the memory networks, we implemented the end-to-end version as described in \cite{WestonMM,DBLP:conf/nips/SukhbaatarSWF15}.
	Of the 634 crowd-sourced labeled examples described in section~\ref{labsoundsources}, we used 100 as test data, the rest as training data.
	Model parameters such as hidden layer size of the memory networks were tuned using cross-validation on the training data. As  shown in Table 2, we obtain high accuracy across all models. The best performing model is a linear model with an LSTM encoding of the sound phrases, achieving accuracy of 90\%. Surprisingly, we could not obtain better results with the memory networks model. Increasing the memory size or the number of hops (how often we iterate over the memories) did not help. One possible reason  is the size of our training data, in previous work \cite{WestonMM,DBLP:conf/nips/SukhbaatarSWF15}, the memory networks were trained on 1,000 or more examples per problem type whereas our training data is half the size. Nevertheless, the memory networks module still produces good accuracy, with best  performance of  87\%.
	
	%Results are shown in Table \ref{tbl:soundsourcerelation}
	%Test on 100, train on 
	%Testing, tested on 100.
	%total other:  534 534 100 100
	
	\begin {table}[t]
	\centering
	\begin{tabular}{|l|l|}
		\hline
		\textbf{Learning Model }& \textbf{Accuracy} \\
		\hline
		LM: LSTM encoder & \textbf{0.90} \\
		LM: (Source - Target) & 0.88 \\
		LM: (Target - Source) &  0.87 \\
		LM: Vector Concatenation &  0.83 \\
		MM: 1 hop & 0.87\\
		%MM: 2 hops &  0.84\\
		MM: 3 hops &  0.85 \\
		%MM: 4 hops &  0.86 \\
		\hline
	\end{tabular}
	\label{tbl:soundsourcerelationtable}
	\caption{Accuracy of the linear models (LM) and memory networks models (MM) on the sound-source relation. }
	\end {table}
	
	\section{Sound-Scene Relationship}
	%The sound-source relationship represents information about which objects produce which sounds. For example that planes and birds are capable of \textit{flying},  the wind  \textit{blows}, and   geckos \textit{bark}. Obtaining sufficient labeled data for this relationship is not straightforward.
	The sound-scene relationship represents information about which sounds are found in  which scenes. For  example, birds chirping can be found in a forest. Therefore,  this kind of information can also be used in context recognition systems \cite{eronen},  in addition to providing common sense knowledge that could be useful  in language understanding tasks.
	
	\paragraph{Labeled Data Generation.}
	%To obtain training data we could solicit input on a crowd-sourcing platform by asking participants to list examples of sounds and their sources. However,  such an approach will be  expensive in terms of the time required for participants to produce enough  examples and hence what we would need to pay the participants. Instead, we propose an  approach  which allows us to specify a crowd-sourcing task which only requires   \textit{``yes/no/notsure"} answers.
	We would like to obtain labeled data in the form of scenes and their sounds. For example,  (beach, waves crashing), (construction, hammering), (street, sirens), (street, honking cars).
	To obtain this type of labeled data, we again would like to only use   \textit{``yes/no/notsure"} crowd-sourcing questions. To generate plausible sound-scene pairs, first we find all sentences that mention at least one scene  and one  sound  concept. To detect sound concepts, we use the approach described in Section~\ref{labsoundsources}. To detect mentions of scenes,  we specified a list of  36  example scenes, which includes scenes such as beach, park, airport most of our scenes are part of  the list of acoustic scenes from a scene classification challenge~\footnote{http://www.cs.tut.fi/sgn/arg/dcase2016/}.  The full  list of scenes is in the supplementary data accompanying this submission. For every sentence that mentions both an acoustic scene and a sound concept, we  apply a dependency parser\footnote{https://pypi.python.org/pypi/practnlptools/1.0}. This step produces  dependencies that form a directed graph, with words being nodes and dependencies being edges.  
%	For example, the sentence: \textit{``The park was filled with the sound of children playing"}, yields the following dependencies:
	
%	{\footnotesize
%		\begin{center}
%			\begin{minipage}[c]{0.8\linewidth}
%				\textit{det(park-2, The-1)}\\
%				\textit{nsubjpass(filled-4, park-2)}\\
%				\textit{auxpass(filled-4, was-3)}\\
%				\textit{root(ROOT-0, filled-4)}\\
%				\textit{det(sound-7, the-6)} \\
%				\textit{nsubj(playing-10, sound-7)}\\
%				\textit{prep\_of(sound-7, children-9)}\\
%				\textit{prepc\_with(filled-4, playing-10)'}
%			\end{minipage}
%		\end{center}
%	}
	
	%The details of the dependency relations can be found in \cite{de2008stanford}. Next, we traverse the dependency graph in order to obtain the path between the mention of a sound concept, in this case ``children playing", and the mention of the acoustic environment ``park". 
	
	Dependency graph shortest paths between entities have been found to be a good indicator of relationships between entities \cite{XuMLCPJ15,NakasholeWS13}.
%	 In our example, the shortest path between \textit{park} and \textit{children playing},  labeled with edge and node names  is as follows: ``nsubjpass() filled prepc\_with() sound prep\_of()".
	We  use shortest paths as features in order classify sound-scene pairs.
	% into those that express the relationship of interest  (SoundFoundInEnvironment) and those that do not. Classifier training would require labeled training data.  
	To obtain training data, we sort the paths by frequency, that is, how often we have seen the path occur with different sound-scene   pairs.  We then consider pairs that occur with frequent shortest paths to 
	%Among the most frequent paths, we label the paths yes or no, depending on whether they express the  relationship of interest. 
	be  plausible sound-scene pairs which we can present to crowd-workers in  \textit{``yes/no/notsure"} questions.
	We  randomly selected   584 sound-scene pairs, and the corresponding sentences that mention them, which were then presented to crowd workers in questions.
	% These sound-scene pairs were presented in a Mechanical Turk task for annotation with  \textit{``yes/no/notsure"} answers, whether or not the pair is valid for the \textit{soundScene} relation.
	%Thus from ``Birds chirping"   we generate the question, ``Is  chirping a sound produced by  birds?"  We generated 634 such questions on which we obtained a moderate  inter-annotator agreement rate of Fleiss $\kappa =0.57$, see the first row of  Table \ref{tbl:soundsoucesMturk}.
	The  inter-annotator agreement rate on this task is Fleiss $\kappa =0.35$,  see Table \ref{tbl:soundsoucesMturk}.
	
	\paragraph{Learning Models and Evaluation.} 
	We use the learning models described in Sections~\ref{linearmodel} and \ref{memorynetworks}. For the linear model, we consider three options for features.
	%\begin{description} 
	\textit{Shortest Paths (SP)}:  LSTM encoding of the dependency shortest path. \textit{Sentence (S)}: an LSTM encoding of the  sentence.
	\textit{ SP + S}: encoding of both the shortest path and the sentence are used as  features.
	For the memory network models, we considered using the contents of both the shortest paths and the sentences to produce memories. 
	We use 100 of the 584 labeled data for testing, the rest for training. The shortest paths performed better, for space reasons we omit the results of using sentences as  memories.  As shown in Table \ref{tbl:soundEnvironmenttrelation}, the linear model with the shortest path achieves the best accuracy of 81\%. However, the best performing memory networks model with 3 memory hops is not significantly worse at 80\% accuracy. 
	
	\begin {table}[t]
	\centering
		%\resizebox{\columnwidth}{!}{%
	\begin{tabular}{|l|l|}
		\hline
		\textbf{Learning Model }& \textbf{Accuracy}\\
		\hline
		LM: shortest path & \textbf{0.81} \\
		LM: shortest path +sentence: &  0.80 \\
		LM: sentence &  0.75 \\
		MM:	1 hop &  0.75 \\
		%MM: 2 hop &  0.78 \\ 
		MM: 3 hops &  0.80 \\
		%MM: 4 hops & 0.77\\
		%	4 hops:  & : 0.75 \\
		%	Using sentences as memories. & \\
		%	1 hops &  0.729 \\
		%	2 hops &  0.71875 \\
		%	3 hops &  0.739 \\
		%	4.hops &  0.697 \\
		\hline
	\end{tabular}

	\label{tbl:soundEnvironmenttrelation}
	\caption{Accuracy  on the sound-scene relation.}
	\end {table}
	
	\begin {table}[t]
	\centering
	\begin{tabular}{|l|l|}
		\hline
		\textbf{Learning Model }& \textbf{Accuracy} \\
		\hline
		LM: LSTM encoder &\textbf{0.84}\\
		LM: vector addition & 0.81 \\
		MM: 1 hop &  0.82 \\
		%MM: 2 hops & 0.82 \\
		MM: 3 hops & 0.82 \\
		%MM: 4 hops & 0.81 \\
		\hline
	\end{tabular}
	\label{tbl:smellsentimentrelation}
	\caption{Accuracy on the sound-sentiment relation.}
	\end {table}
	
	\section{Smell-Sentiment Relationship}
%	The sound-source relationship represents information about which objects produce which sounds. For example that planes and birds are capable of \textit{flying},  the wind  \textit{blows}, and   geckos \textit{bark}. Obtaining sufficient labeled data for this relationship is not straightforward. 
	For the  smell-sentiment relationship, the goal is to extract information about  which smells are considered  pleasant, unpleasant or neutral. In general, sentiment is both subjective and context dependent. However,  as we show through crowd-sourced  annotations, there is substantial consensus even on sentiment of smells.
	% The same smell can have different sentiments across different contexts, we learn to extract the relationship within a given context. Additionally, there are smells whose sentiment remains the same regardless of context.
	
	\paragraph{Labeled Data Generation.}
	
%	We would like to obtain labeled data in the form of scenes and their sounds. For example,  (beach, waves crashing), (construction, hammering), (street, sirens), (street, honking cars).
%	To obtain this type of labeled data, we again would like to only use   \textit{``yes/no/notsure"} crowd-sourcing questions. To generate plausible sound-scene pairs, first we find all sentences that mention at least one scene  and one  sound  concept. To detect sound concepts, we use the approach described in Section~\ref{labsoundsources}. To detect mentions of scenes,  we specified a list of  36  example scenes, which includes scenes such as beach, park, airport most of our scenes are contained in  the list of acoustic scenes from a scene classification challenge~\footnote{http://www.cs.tut.fi/sgn/arg/dcase2016/}.  The full  list of scenes is part of supplementary data accompanying this submission. For every sentence that mentions both an acoustic scene and a sound concept, we  apply a dependency parser\footnote{https://pypi.python.org/pypi/practnlptools/1.0}. This step produces  dependencies that form a directed graph, with words being nodes and dependencies being edges.  
	
%	To generate labeled data for the smell-sentiment relationship, we again seek to present a highly simplified,  multiple choice crowd-sourcing task 
	% requiring participants to choosing between one of $k$ options, where $k$ is a small number. 
	%Here we do not pose a task requiring   \textit{``yes/no/notsure"} answers. Instead we pose a task requiring the answers of  \textit{``pleasant/unpleasant/neutral/notsure"}.
	First we generate a list of plausible  smells, following a similar approach  to Section~\ref{labsoundsources}. That is, we  search for  the pattern: ``\text{ smell} \text{ of } \text{ \textless y\textgreater}" in the ClueWeb corpus. The result  is a large  collection of  occurrences such as:
	%\begin{eqnarray*}
	``\text{ smell} \text{ of \textit { rotten eggs.}}" or \text{ smell} \text{ of \textit { cherry blossoms.}}
	From this collection, we randomly selected a sample of  500  phrases and asked 3 crowd workers per  phrase on Mechanical Turk,  to say \textit{``yes/no/notsure"} if they agree the  phrase refers to a smell concept.
	%  From this annotation, we consider a noun phrase to be a true mention of smell concept if the majority of the participants chose ``yes". 
	By the majority vote measure metric  $89.9\%$ of the 500  phrases  are true mentions of smells, with an somewhat low agreement rate of 0.33 Fleiss $\kappa$.
	% Having quantitatively verified these noun phrase collection contains a substantial fraction of noun phrases that refer to smell concepts,
	Having verified that our list of  phrases contains a substantial number of smell concepts, we then use these  phrases  to evaluate sentiment of smells in a different Mechanical Turk task. We present a phrase within a sentence context. We then asked crowd workers to choose if  the phrase refers to a smell that is  \textit{``pleasant/unpleasant/neutral/notsure/notasmell"}.  We generated 600 such questions on which we obtained a moderate  inter-annotator agreement rate of Fleiss $\kappa =0.43$, see Table \ref{tbl:soundsoucesMturk}. While this is not a yes/no task, it is still a simple multiple choice task with the same  advantages of the yes/no tasks as we described earlier.
	
	%\subsection{Learning model}

	\paragraph{Learning Models and Evaluation.} 
		We again  use the learning models described in Sections~\ref{linearmodel} and \ref{memorynetworks}. For the linear model, we consider two options for features.
		%\begin{description} 
		\textbf{LSTM encoder}:  LSTM encoding of the smell phrase
		\textbf{Vector addition}: vector addition encoding of the smell phrase. 
%		It is worth noting that  we only consider phrases of length  up to 2 words.
		For the memory network models, the contents of the sentence that mentions the phrases are stored as memories. We use 100 of the 600 labeled data for testing, the rest for training. As can be seen in Table~\ref{tbl:smellsentimentrelation}, the linear model with LSTM encoded phrases  achieved the highest accuracy of 84\%.

	%\subsection{}
%	\paragraph{Learning models and Results.}
%	We can then use the data to train models. Here we again use the linear model and the memory network model.  We used ?? for training and ?? for testing. The accuracy numbers are shown.
%	Linear model and memory network again. Features are word embeddings.
%	
%	Results are in \ref{tbl:smellsentimentrelation}.
%	Ask turkers, 600 examples. Tested on 100.
%	%600 0.88 528
	
%\section{Discussion}
	
	%What does it mean for future common sense relationships.
	
\section{Conclusion}
%Lack of training data is one of the factors impeding progress in commonsense knowledge acquisitions.

Cyc~\cite{Lenat95},  and  ConceptNet~\cite{havasi2007conceptnet} are well-known examples of attempts to build  knowledge bases of everyday common sense knowledge. These projects are decades long  manual efforts involving either experts or crowd-sourcing.  Other knowledge bases  focus on facts about named entities  such as people, locations, and companies \cite{Bollacker2008,hoffartetal2012a,MitchellCHTBCMG15}.
% Common sense  contained in these knowledge bases is limited to part-whole relations, and taxonomic  relations

 In this paper,  we extracted novel common sense relations. To obtain labeled data, we proposed a combination of large corpora, and  multiple choice  crowd-sourced questions. These type of questions require little effort from crowd workers while mitigating the amount of noise one might get from open-ended  questions.
We have also proposed and trained models on this data, achieving high accuracy for all relations.
Scaling up our  approach to more relations is  an exciting future direction for our work. 
 We believe our   technique can scale given its minimally-supervised nature. 
%Here for each relation we can up with one pattern that is high yield but potentially noisy an used crowd workers for filtering. The pattern aspect could also be crowd-sourced by soliciting a few patterns from crowd workers. We believe this is a promising direction towards building true common sense knowledge bases to compliment today's mostly named-entity centric knowledge bases. 
%Scaling up our  approach to more relations is therefore a naturally exciting future direction for our work. 

%While we have tested our approach on few relations, the implications of our results are useful. Our results show that we can indeed extract more common sense relations toward making knowledge bases more useful for language understanding and other artificial intelligence tasks that require common sense knowledge. 

	\clearpage
	\nocite{*}
	\bibliographystyle{acl2016}
	\bibliography{sound}

\end{document}